%% file: main.tex
\documentclass[11pt]{article}
\usepackage[preprint]{acl}
\usepackage{times}
\usepackage{latexsym}
\usepackage[T1]{fontenc}
\usepackage[utf8]{inputenc}
\usepackage{microtype}
\usepackage{graphicx}
\usepackage{booktabs}
\usepackage{amsmath}
\usepackage{amssymb}
\usepackage{placeins}

\title{Geometric Representations of African Languages: \\ A Regional Semantic Hub and Cultural Steering}
\input{authors}

\begin{document}
\maketitle

\begin{abstract}
We study how Gemma 4 31B represents African languages and responds to cultural steering. The first study compares nine African languages and three controls using probes, contrast directions, and measures of representation similarity. Transfer from English varies across languages and layers. Directions representing an Africa versus West contrast are more aligned among the African languages than between these languages and the controls at several layers. The comparison across language families passes the reported Holm threshold at five of twelve layers, although dependence between language pairs limits the statistical interpretation. Within Nigeria, Yoruba and Igbo are more aligned than the average of their pairs with Hausa at eleven of twelve layers. The second study uses separate English data to construct directions for Nigeria, Ghana, Kenya, and South Africa. Under union scoring at the selected strengths, estimated differences in attribution rates from random directions range from 0.63 to 0.81. Most outputs pass the automated structural coherence screen. Comparisons with Aya Expanse 32B show that results depend on the representation measure. Together, the studies document regional and family patterns in the sampled representations and country steering in English.
\end{abstract}

\section{Introduction}
\label{sec:intro}
African languages remain underrepresented in language model development and evaluation \citep{nekoto2020participatory, adelani2021masakhaner}. Work on multilingual representations asks whether models process different languages through an English oriented internal representation \citep{wendler2024llamas} or through a shared semantic space that is not specific to one language \citep{wu2024semantichub}. Studying African languages broadens the language coverage of this question and provides comparisons across language families within a region.

We examine two aspects of internal representation in a frozen multilingual model. Study~1 measures cross language similarity in a translated corpus labelled by entity type and by an Africa versus West content contrast. It asks whether information transfers from English and whether contrast directions are more similar among African languages than to non African controls. Study~2 constructs country versus Western directions from a separate English corpus and tests whether adding these directions during generation elicits content associated with Nigeria, Ghana, Kenya, or South Africa.

In this paper, a regional semantic hub refers to greater alignment of the sampled African language directions on the chosen content contrast. This operational description does not establish a general semantic space shared by all African languages. Both studies examine representations of African languages and cultural content in the same model. Study~1 compares how a regional contrast is represented across languages. Study~2 tests whether directions derived from country content can alter English generation. Each study constructs its own directions, and a separate pilot selects the steering layer. The experiments therefore answer separate questions about representation and intervention; they do not test whether the regional pattern causes the steering effect.

Our contributions are:
\begin{itemize}
\item A comparison of nine African languages and three controls using four measures of internal representation, including a comparison within Nigeria that holds country fixed while comparing pairs from the same family and from different families.
\item Evidence that Africa versus West directions are more aligned among the African languages than with the controls at several layers, with the strongest separation at layers 14 to 24. The comparison across families meets the reported Holm threshold at five of twelve layers; the result within Afro Asiatic remains inconclusive.
\item An evaluation of four country specific steering directions in English against random directions of matched norm, including specificity, coherence, and extraction method comparisons.
\item A comparison of English reference measures with Aya Expanse 32B, showing that the model ordering depends on the measure.
\end{itemize}

\section{Related Work}
\label{sec:related}
\paragraph{Multilingual representations.}
Cross language transfer has motivated analyses of shared representations \citep{pires2019multilingual, conneau2020unsupervised}. \citet{wendler2024llamas} investigate the latent language of multilingual transformers using the logit lens, which maps intermediate hidden states through the output vocabulary. \citet{wu2024semantichub} study shared semantic representations across languages and modalities. Our measurements compare hidden states directly. They measure similarity and transfer on the selected corpus. The language used during token prediction is outside the scope of these measurements.

\paragraph{African languages and language coverage.}
Participatory machine translation and African language named entity recognition have documented the need for broader language coverage and locally grounded resources \citep{nekoto2020participatory, adelani2021masakhaner}. We analyse internal states to examine the information represented by the model. A successful probe shows that the selected information can be recovered from those states; it does not by itself explain downstream failures or establish that a model uses the recovered information during generation.

\paragraph{Comparing representations.}
Linear probes measure information recoverable from frozen activations \citep{alain2018probes}. SVCCA compares correlated low dimensional representations \citep{raghu2017svcca}, whereas orthogonal Procrustes measures agreement under an orthogonal transformation. Different similarity measures do not rank representations identically \citep{kornblith2019similarity}. We report each measure separately because they assess different properties. Directions constructed from differences in means compare a specific labelled contrast \citep{park2024linear}.

\paragraph{Cultural steering.}
Activation addition and contrastive activation addition modify generation by adding directions to internal states \citep{turner2024actadd, rimsky2024caa}; related work studies single direction behavioural interventions \citep{arditi2024refusal}. Cultural steering has been investigated through localized knowledge, cultural values, and country specific generation \citep{veselovsky2025localized, dang2026cultural, khanuja2026steering}. Studies of cultural alignment also motivate examining which groups a model's outputs represent \citep{tao2024cultural, alkhamissi2024investigating}. We evaluate a difference in means approach for four African countries. Our random direction comparison tests whether the selected direction matters.

\section{Models and Data}
\label{sec:setup}
The primary model is Gemma 4 31B, instruction tuned \citep{gemmateam2026gemma4}, recorded as \texttt{google/gemma-4-31B-it}, with 60 layers and 5376 hidden dimensions. Measurements use its twelve sampled layers, indexed from zero as L4, L9, $\ldots$, L59. The secondary model is Aya Expanse 32B \citep{cohere2024aya}, recorded as \texttt{CohereForAI/aya-expanse-32b}, with 40 layers and 8192 hidden dimension; measurements use L0, L4, $\ldots$, L36, L39. Both models use bfloat16 precision. Aya is evaluated on Swahili, Hausa, Amharic, Zulu, Yoruba, and Finnish. The comparison between models using English as a reference uses the five shared African languages. An additional saved pairwise comparison uses these five languages and Finnish as its sole control (Appendix~\ref{app:aya}). Steering uses Gemma alone.

\paragraph{Study 1 corpus.}
Each record pairs a full English sentence with a translation generated using Gemini 3 Pro \citep{googledeepmind2025gemini}. Each of the twelve language files contains 1{,}628 pairs: 37 templates applied to 44 entities, with 814 Africa labelled and 814 West labelled records. The entity classes contain 444 city, 444 language, 370 food, and 370 landmark records. The twelve files share the same 1{,}628 English sentences and entity identities. Entity type defines the four class probe task; region defines the difference in means direction. These are different prediction targets. Appendix~\ref{app:corpus} gives the entity counts and a parallel example. Table~\ref{tab:languages} gives the language groups. Danish and Finnish provide European controls, and Quechua provides a low resource non African comparison. These controls broaden coverage but do not establish that training resources or translation quality are matched across languages.

\begin{table}[t]
\centering\small
\begin{tabular}{ll}
\toprule
Group & Languages \\
\midrule
Niger Congo & Swahili, Yoruba, Igbo, Zulu, Wolof \\
Afro Asiatic & Hausa, Amharic, Somali \\
Other African & Songhai \\
Controls & Danish, Finnish, Quechua \\
\bottomrule
\end{tabular}
\caption{Language groups used in the comparisons. Songhai contributes to the all African comparison but not the two family level comparisons.}
\label{tab:languages}
\end{table}

\paragraph{Study 2 data.}
Steering uses a separate set of 800 English contrastive pairs, with 200 per country. Each pair contains a sentence about a target country's culture and a Western counterpart. Evaluation uses 40 open ended prompts that do not name a target country: 37 labelled cultural prompts and three labelled controls in rubric version 2.0. The saved aggregate rates include these controls. Appendix~\ref{app:data} gives a construction example and documents the implemented scoring exclusions. This construction dataset is separate from the multilingual corpus in Study~1.

\paragraph{Translation and evaluation provenance.}
Gemini 3 Pro was used for Study~1 translations, and Gemini 3 Pro Preview evaluates Study~2 generations. The instruction tuned Gemma 4 31B model is the primary model whose representations we analyse. Translation fidelity is not established by an independent human assessment in the reported evaluation. The translator, judge, and primary model also share a developer. Translation and cultural content may both affect the measured representations. Our conclusions apply to this corpus and language sample.

\paragraph{Compute.}
Experiments used a single A100 80GB instance on Google Cloud, at a reported total cost of approximately USD 1{,}000. This includes activation extraction, the Study~1 measures, and the Study~2 sweep over three extraction methods, four country and five random vectors, six strengths, 40 prompts, and five samples per prompt.

\section{Study 1: Methods}
\label{sec:study1methods}
\paragraph{Sentence representations.}
For sentence $x$, let $h_t^{(\ell)}(x)\in\mathbb{R}^d$ be the output of transformer block $\ell$ at token position $t$. We tokenize raw sentences without chat formatting, using padded batches of 16 and truncation at 128 tokens. We average over positions with attention mask value one, denoted $T_x$:
\begin{equation}
\bar h^{(\ell)}(x)=\frac{1}{|T_x|}\sum_{t\in T_x}h_t^{(\ell)}(x).
\end{equation}
Each sentence thus contributes one fixed length vector per layer. Padding is excluded; special tokens added by the tokenizer are not separately removed. Pooling uses float32 values captured by a forward hook on the selected block's output. The four measures below use these pooled representations. 

\paragraph{Entity type probe.}
A four class linear classifier is trained on frozen representations using Adam, learning rate 0.01, for 100 epochs. We report English to English, target to target, and English to target accuracy; the last applies the English classifier to target language representations without retraining. Uniform guessing gives 25\%; the full corpus's majority class baseline is 27.3\%. The variant with shared entities randomly splits records 75:25 for training and evaluation, so entity identities can occur in both. For evaluation on unseen entities, each of five folds holds out one African and one Western entity per class, trains on the remaining entities, and evaluates on all records for the eight excluded entities. This assesses generalization beyond the training entities, while retaining the corpus's templates and construction process. The majority class baseline depends on the class counts in each test fold.

\paragraph{Africa versus West direction.}
For the Africa labelled set $A$ and West labelled set $W$ in each language, define
\begin{equation}
\begin{split}
\mu_A&=\frac{1}{|A|}\sum_{x\in A}\bar h(x),\qquad
\mu_W=\frac{1}{|W|}\sum_{x\in W}\bar h(x),\\
v&=\frac{\mu_A-\mu_W}{\|\mu_A-\mu_W\|_2}.
\end{split}
\label{eq:direction}
\end{equation}
The English to target angle is $\theta=\arccos(v_{\mathrm{EN}}^\top v_{\mathrm{XX}})$, reported in degrees. A smaller angle indicates closer alignment of this specific contrast; it does not establish equivalence of the complete representation spaces. Directions, SVCCA, and Procrustes are computed over ten seeded 80\% subsamples drawn without replacement. Reported metric summaries average these ten runs.

\paragraph{SVCCA.}
Singular Vector Canonical Correlation Analysis compares correlated components of the English and target representations. Each side is reduced by principal component analysis to rank $k\in\{20,40\}$. We average the top five canonical correlations and subtract the corresponding mean over ten runs with shuffled pairs:
\begin{equation}
S=\frac{1}{5}\sum_{i=1}^{5}\rho_i-
\frac{1}{10}\sum_{s=1}^{10}\left(\frac{1}{5}\sum_{i=1}^{5}\rho_i^{(s)}\right).
\end{equation}
Here $\rho_i$ denotes a canonical correlation and $\rho_i^{(s)}$ its counterpart in a shuffled run. A positive value indicates correlation above this shuffled baseline.

\paragraph{Orthogonal Procrustes.}
Let $X$ and $Y$ contain representations of paired English and target sentences, with each set centred on its mean. We fit an orthogonal transformation and report a normalized residual similarity:
\begin{equation}
\begin{split}
R^*&=\mathop{\arg\min}_{R^\top R=I}\|XR-Y\|_F,\\
P&=\operatorname{clip}_{[0,1]}\left(1-
\frac{\|XR^*-Y\|_F^2}{\|X\|_F^2+\|Y\|_F^2}\right).
\end{split}
\end{equation}
The Frobenius norm $\|\cdot\|_F$ summarizes the matrix residual. Larger $P$ indicates closer alignment.

\paragraph{Pairwise regional and family comparisons.}
At each sampled layer, we compare language directions normalized to unit length within each matched subsampling seed and average their cosines: $M_{ab}=\frac{1}{10}\sum_{s=0}^{9}v_{a,s}^{\top}v_{b,s}$. We compare cosines within all African languages, within Niger Congo, and within Afro Asiatic with the corresponding cosines to controls. A fourth comparison tests Niger Congo to Afro Asiatic pairs against Niger Congo to control pairs. One sided Mann Whitney tests summarize whether the first set tends to have larger values. Holm correction adjusts the tests across layers within each comparison: twelve for Gemma and eleven for Aya. Appendix~\ref{app:pairwise} reports Gemma means and adjusted values; Appendix~\ref{app:aya} gives the saved Aya comparison.

Language pairs reuse the same directions, so their observations are dependent. Holm adjustment accounts for the number of layer comparisons. Dependence between pairs remains, so we use the test results as descriptive evidence. Within Nigeria, we compare the cosine for Yoruba and Igbo with the mean of the cosines for Yoruba with Hausa and Igbo with Hausa. This holds country fixed, but it does not control all differences between languages. Tests and bootstrap intervals across layers are also limited by dependence between layers.

\paragraph{Cross model comparison.}
We compare the English reference measures at relative depths from 10\% to 90\%, choosing each model's nearest sampled layer. For each measure and depth, we summarize the five paired African language differences by their median and a 95\% bias corrected and accelerated bootstrap interval from 10{,}000 resamples. Positive differences favour Aya: Gemma minus Aya for the angle, and Aya minus Gemma for the other measures. The probe ratio is English to target accuracy divided by target to target accuracy. The additional Aya pairwise analysis uses fewer African languages and only Finnish as a control. Its coverage differs from the Gemma analysis.

\section{Study 1: Results}
\label{sec:study1results}
\subsection{English reference similarity varies by language and layer}
\label{sec:enhub}
Table~\ref{tab:english} gives the numerical results for all twelve languages. The probe evaluated on unseen entities exceeds 25\% chance at early layers for most languages, with substantial variation in the margin above chance. At L19, accuracy is 71\% for Danish, 61\% for Swahili, 54\% for Amharic, and 26\% for Yoruba. At L59, these values are 32\%, 28\%, 29\%, and 27\%. Holding out entities reduces the stronger transfer scores, while preserving the broad depth pattern (Appendix~\ref{app:probe}).

\begin{table*}[t]
\centering\small
\begin{tabular}{lrrrrrr}
\toprule
& \multicolumn{3}{c}{Accuracy on unseen entities (\%)} & \multicolumn{3}{c}{English reference geometry at L19} \\
Language & L4 & L19 & L59 & Angle ($^\circ$) & SVCCA ($k=40$) & Procrustes \\
\midrule
Swahili & 48 & 61 & 28 & 75 & 0.69 & 0.77 \\
Yoruba & 29 & 26 & 27 & 76 & 0.68 & 0.60 \\
Igbo & 32 & 33 & 26 & 69 & 0.68 & 0.63 \\
Zulu & 33 & 40 & 27 & 84 & 0.68 & 0.63 \\
Wolof & 31 & 38 & 26 & 50 & 0.66 & 0.58 \\
Hausa & 41 & 44 & 29 & 69 & 0.69 & 0.68 \\
Amharic & 50 & 54 & 29 & 54 & 0.69 & 0.75 \\
Somali & 35 & 49 & 27 & 72 & 0.68 & 0.67 \\
Songhai & 36 & 38 & 27 & 52 & 0.64 & 0.56 \\
\midrule
Quechua & 37 & 51 & 26 & 51 & 0.62 & 0.50 \\
Finnish & 34 & 55 & 26 & 42 & 0.68 & 0.69 \\
Danish & 54 & 71 & 32 & 26 & 0.69 & 0.79 \\
\bottomrule
\end{tabular}
\caption{Gemma results at the precision shown in the heatmaps. Probe chance is 25\%. L4, L19, and L59 illustrate three selected depths; the L19 values use the same layer for every language. Angles compare Africa versus West directions, whereas the probe predicts entity type. Higher SVCCA and Procrustes scores indicate greater similarity under their respective definitions.}
\label{tab:english}
\end{table*}

\begin{figure}[t]
\centering
\includegraphics[width=\columnwidth]{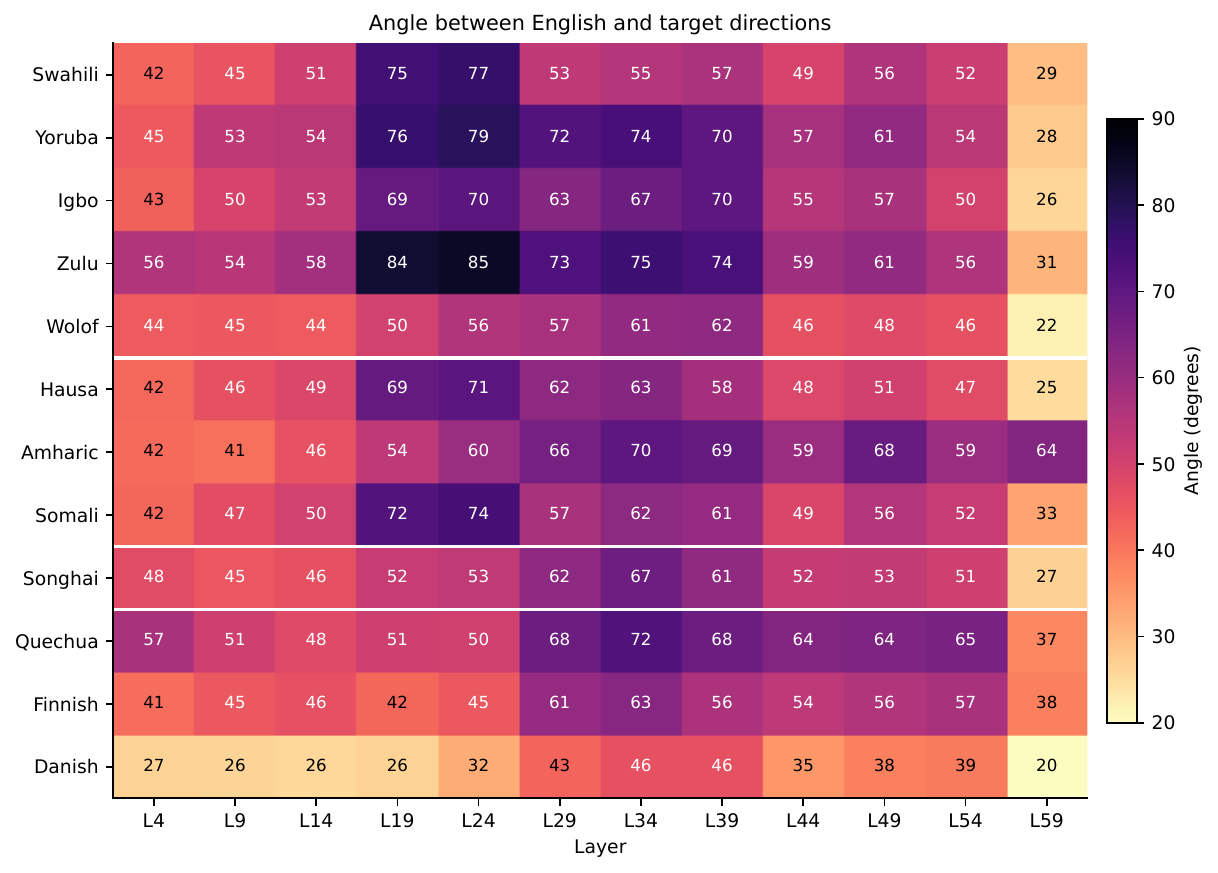}
\caption{Angles between English and target language Africa versus West directions across Gemma layers. Smaller angles indicate closer alignment.}
\label{fig:dimangle}
\end{figure}

The angle between the Africa versus West directions generally increases away from the input and decreases toward the output (Figure~\ref{fig:dimangle}). Six African languages reach their largest angle at L24; Quechua and Finnish peak at L34. Amharic remains relatively rotated from English at the final layer.

SVCCA varies less across languages than the probe: at L19 its scores range from 0.62 for Quechua to 0.69 for several languages. Procrustes distinguishes these representations more strongly, ranging from 0.50 for Quechua to 0.79 for Danish at that layer. African languages do not form a consistently separate group from the controls across these English reference measures. Resource availability is one possible explanation for the language ordering, but the design does not separate it from tokenization and translation quality. We therefore do not interpret the ordering as a measured training resource effect.

\subsection{Regional alignment is strongest at early and middle depth}
\label{sec:africanhub}
\begin{figure*}[t]
\centering
\includegraphics[width=\textwidth]{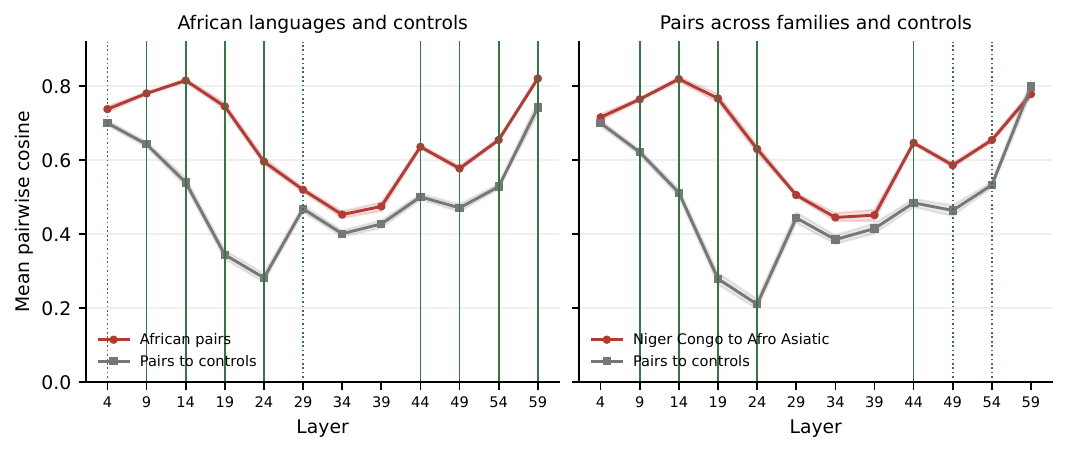}
\caption{Mean cosines among African languages and across the two language families, compared with control pairs. Solid green lines mark Holm adjusted $p<0.05$; dotted green lines mark only raw $p<0.05$. Bands reproduce the saved bootstrap intervals over direction subsamples. The tests reuse languages, which limits their statistical interpretation.}
\label{fig:pairwise}
\end{figure*}

The pairwise comparison shows greater alignment among African language directions than between these directions and the controls at several layers (Figure~\ref{fig:pairwise}). At L19 the all African mean cosine is 0.746, compared with 0.345 for African to control pairs. The corresponding means are 0.815 versus 0.540 at L14 and 0.596 versus 0.282 at L24. Table~\ref{tab:contrasts} summarizes the reported corrected thresholds, subject to the dependence limitation in Section~\ref{sec:study1methods}.

\begin{table}[t]
\centering\small
\begin{tabular}{lc}
\toprule
Comparison & Layers with adjusted $p<0.05$ \\
\midrule
Niger Congo vs control & 8/12 \\
All African vs control & 8/12 \\
NC to AA vs NC to control & 5/12 \\
Afro Asiatic vs control & 0/12 \\
\bottomrule
\end{tabular}
\caption{Reported Mann Whitney comparisons, Holm adjusted across twelve layers within each comparison. NC: Niger Congo; AA: Afro Asiatic. Shared languages make the pairwise observations dependent.}
\label{tab:contrasts}
\end{table}

The comparison across families meets the reported threshold at L9, L14, L19, L24, and L44. The observed alignment therefore crosses the family boundary at these layers. The comparison within Afro Asiatic meets the threshold at no layer. With only three languages in that group, its result remains inconclusive.

Within Nigeria, the Yoruba and Igbo pair is more aligned than the mean of the two Hausa pairs at eleven of twelve layers, with an average difference of 0.116 and a reported 95\% bootstrap interval across layers of [0.078, 0.151]. The original paired test gives $t(11)=5.93$ with one sided $p<0.001$; sign and Wilcoxon checks give the same direction of evidence. These results across layers are descriptive because the layers are dependent. At L19 the family gap is approximately 0.08, compared with the all African to control gap of 0.40. These two gaps describe L19 only.

The late layer regional differences are smaller. At L59, within African and African to control means are 0.821 and 0.742. Quechua's position outside the African grouping shows that the pattern is not shared by every low resource language in this sample; it does not rule out translation, tokenization, or content differences as explanations.

\subsection{Cross model differences depend on the measure}
\label{sec:crossmodel}
Aya does not reproduce every Gemma depth pattern. Its probe transfer for Finnish and Hausa peaks later, and its angle map does not show the same overall shape. Across the five African languages, Aya has higher SVCCA and Gemma has higher Procrustes alignment at all nine matched depths, with reported bootstrap intervals excluding zero. Aya also has higher probe transfer at depths from 50\% to 90\%, with intervals excluding zero. Its probe ratio is higher at these depths, but the interval at 70\% includes zero, $[-0.0029, 0.3188]$. The probe ratio intervals at 50\%, 60\%, 80\%, and 90\% exclude zero. Most angle differences remain inconclusive. In the additional saved Aya pairwise analysis, the all African comparison meets the reported Holm threshold at L32 and L36 (2/11 layers), while neither within family comparison nor the comparison across families does so at any layer (Appendix~\ref{app:aya}). This smaller sample provides limited evidence about whether the Gemma pattern extends to Aya.

\section{Study 2: Methods}
\label{sec:study2methods}
For each country $c$, we compute a unit direction $v_c=(\mu_c-\mu_W)/\|\mu_c-\mu_W\|_2$ from English country and Western anchor sentences. During generation we add it to the residual stream, the model state passed between layers, using $h\leftarrow h+\alpha v_c$. Extraction and injection use L24, selected by a pilot using Nigeria alone over layers and strengths. This choice was not derived from Study~1's regional analysis.

The primary extractor, \texttt{chat\_mean}, averages all nonpadding token representations from inputs formatted with the chat template, including the instruction and role/generation markers; it does not pool generated response tokens. The user message is ``Complete this sentence in one sentence: [sentence]'', formatted with a generation prompt and thinking disabled. Extraction truncates at 128 tokens and uses batches of 16. We compare it with raw text mean pooling (\texttt{raw\_mean}) and the saved chat last position extractor (\texttt{chat\_last}). The two chat based methods perform similarly at their selected strengths; raw text extraction performs substantially worse (Appendix~\ref{app:extraction}). The intervention is added to the output of block 24 at every position processed by that block during generation, including the initial prompt pass.

\paragraph{Generation and controls.}
We sweep $\alpha\in\{0,20,30,40,50,60\}$, using temperature 0.7, top $p$ 0.9, and a maximum of 80 new tokens. Each condition uses 40 prompts and five samples per prompt, giving 200 generated responses before scoring exclusions. The prompts are wrapped as user turns using the same sentence completion instruction and chat template. Five random unit vectors, with seeds 42 to 46, are injected at the same layer and strengths and pooled at evaluation. The $\alpha=0$ condition is unsteered. Neither control is an explicit country prompting baseline.

\paragraph{Scoring.}
A lexical rubric checks for country specific strict tokens using substring matching that ignores letter case. Gemini 3 Pro receives the prompt, rubric, rubric verdict, and response and returns country attribution and a coherence judgement at temperature zero. A separate structural coherence diagnostic requires at least five words separated by whitespace and a repetition fraction below 0.6, where repetition is one minus the proportion of distinct lowercased words. This diagnostic detects short or repetitive outputs. It provides a limited check of linguistic quality.

We report four scoring modes: rubric only, judge only, intersection (both attribute the country), and union (either attributes it). Union is the primary reported mode. The non rubric modes exclude outputs marked incoherent by the model judge or lacking a country attribution verdict; the structural diagnostic is reported separately and does not filter these rates. Rubric only scoring does not apply the judge coherence filter. The three rubric designated control prompts are included in the saved aggregates, and the implemented diagonal exclusion is the South African northern festival prompt only. Intersection is more restrictive than union on their common eligible outputs.

\paragraph{Rates, uncertainty, and selection.}
The diagonal hit rate is the fraction of eligible outputs attributed to the country corresponding to the steering vector. Off diagonal rates measure attribution to the other countries. Rates have 95\% Wilson intervals. Prompt bootstrap intervals use 10{,}000 resamples of per prompt means, weighting prompts equally. When coherence filtering leaves unequal numbers of responses per prompt, this estimand differs from the response pooled rate underlying the Wilson interval. Both are reported in Appendix~\ref{app:intervals}.

At each country's selected positive strength, we estimate the difference from the pooled random direction baseline using independent Jeffreys posteriors:
\begin{equation}
\begin{split}
p_c&\sim\mathrm{Beta}(k_c+\tfrac12,n_c-k_c+\tfrac12),\\
p_r&\sim\mathrm{Beta}(k_r+\tfrac12,n_r-k_r+\tfrac12).
\end{split}
\end{equation}
Here $k$ and $n$ are hit and eligible trial counts. We report the posterior mean of $p_c-p_r$ and a 95\% credible interval from 20{,}000 draws. The selected strength maximizes the observed diagonal rate, breaking ties toward smaller strength. Selecting the strength on the evaluation data can overestimate both the rate and its difference from random. We therefore report the full strength sweep alongside the selected results.

\section{Study 2: Results}
\label{sec:study2results}
\subsection{Country directions increase attribution to the target country}
For the prompt ``The national football team of this country is affectionately known by the nickname'', an unsteered output names Australia's Socceroos. At $\alpha=50$, the four country directions elicit Super Eagles, Black Stars, Harambee Stars, and Bafana Bafana, respectively; a random direction produces England's Three Lions. This example illustrates the intervention on one prompt. Aggregate results appear in Table~\ref{tab:steering}.

\begin{table*}[t]
\centering\small
\begin{tabular}{lrrrrl}
\toprule
Country & Best $\alpha$ & Union hit rate & Random & Intersection & Difference from random [95\% CrI] \\
\midrule
Nigeria & 40 & 0.89 & 0.08 & 0.62 & 0.81 [0.76, 0.86] \\
South Africa & 40 & 0.79 & 0.08 & 0.55 & 0.71 [0.65, 0.77] \\
Ghana & 50 & 0.71 & 0.04 & 0.44 & 0.66 [0.60, 0.73] \\
Kenya & 50 & 0.68 & 0.05 & 0.54 & 0.63 [0.56, 0.69] \\
\bottomrule
\end{tabular}
\caption{Country steering results with \texttt{chat\_mean} extraction. Union rates use the listed selected strengths; random rates pool five random directions at the same strength. Intersection rates use their own selected strengths: 50 for Nigeria, Ghana, and Kenya; 40 for South Africa. The effect column reports the posterior mean of the difference from random. CrI denotes a posterior credible interval. Selecting the best strength affects both rates and differences.}
\label{tab:steering}
\end{table*}

Under union scoring, differences from random range from 0.63 to 0.81, with the reported intervals excluding zero. The more restrictive intersection rates are lower: 0.62 for Nigeria, 0.55 for South Africa, 0.44 for Ghana, and 0.54 for Kenya. The ordering of Ghana and Kenya thus depends on the scoring rule. At a common strength of 50, union rates are 0.88, 0.71, 0.68, and 0.72 for Nigeria, Ghana, Kenya, and South Africa, respectively, compared with approximately 0.03 to 0.08 for individual random directions. These results show that the directions alter the country content measured by this evaluation.

\subsection{Specificity and coherence vary with strength}
Each country direction eventually elicits its own country more often than the others, but separation is not immediate (Appendix~\ref{app:specificity}). At $\alpha=30$, the Ghana vector produces Nigerian and Ghanaian attribution at similar rates, 39.1\% and 38.6\%. At the common strength $\alpha=50$, the largest confusions are between these countries: the Ghana vector produces Nigerian attribution in 15.0\% of eligible responses, and the Nigeria vector produces Ghanaian attribution in 10.5\%. Shared cultural content is a possible explanation, but this experiment does not isolate its cause.

\begin{figure*}[t]
\centering
\includegraphics[width=\textwidth]{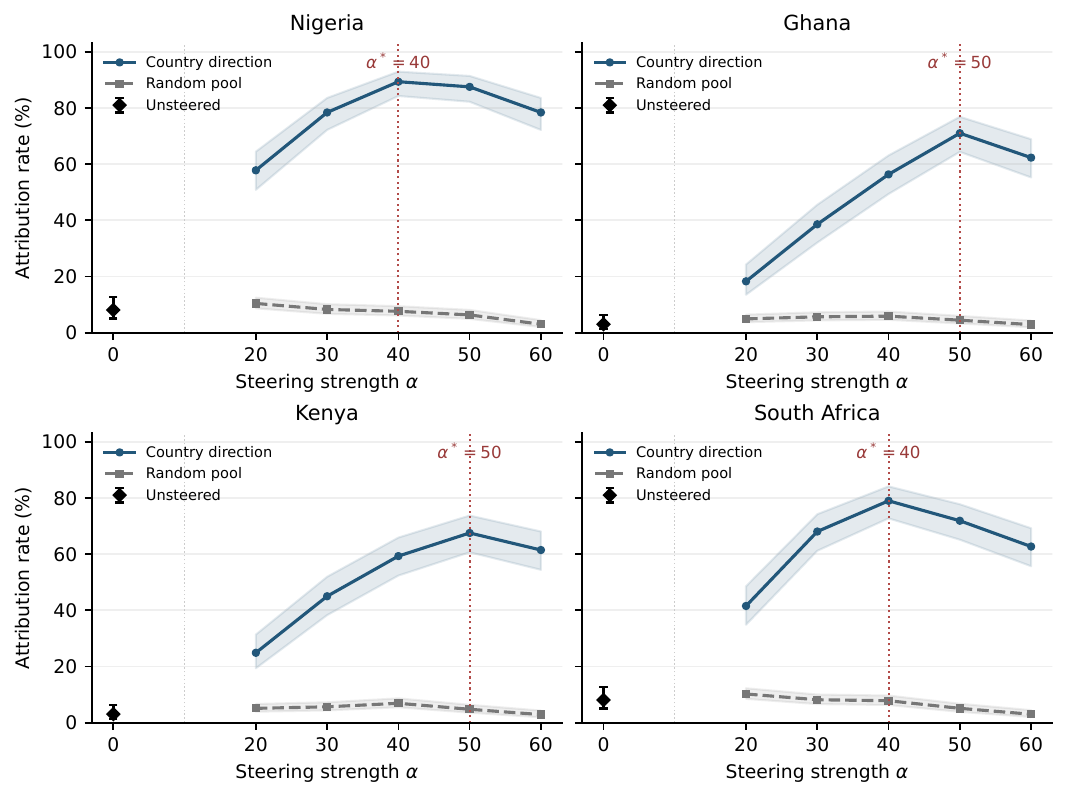}
\caption{Attribution to each target country across steering strengths. Lines show country directions and pooled random directions at positive strengths; bands are the saved 95\% Wilson intervals. The separate point at zero shows the measured unsteered rate over all 40 prompts. South Africa's country curve excludes one prompt, while its unsteered point and random pool retain it. Selected strengths are marked. Appendix~\ref{app:intervals} gives intervals based on prompt means.}
\label{fig:dose}
\end{figure*}

Attribution rises and then plateaus or declines, with selected strengths between 40 and 50 (Figure~\ref{fig:dose}). The attribution rate for each country direction exceeds its random pool at every tested positive strength. The structural coherence rate for pooled country vector outputs is 100\% through strength 50 and approximately 97\% at 60. Random directions reach approximately 85\% pooled coherence at 50, including one seed at 56.5\%. The structural screen and language model coherence judgement agree on 94.1\% of 9{,}200 scored generations. Their agreement is not an independent human validation of coherence.

At high strengths, attribution to both the target and other countries declines while measured coherence remains high. The outputs therefore contain fewer of the country signals recognised by the evaluation.

\section{Discussion}
\label{sec:discussion}
The regional pattern concerns directions representing the Africa versus West content contrast. The largest separation from controls occurs in the early and middle layers. The comparison across Niger Congo and Afro Asiatic meets the reported Holm threshold at five of twelve layers. Within Nigeria, Yoruba and Igbo are more aligned than the average of their pairs with Hausa at eleven of twelve layers. Regional and family patterns therefore coexist in this sample. Shared languages and dependent layers limit the statistical interpretation, while translation and content effects remain possible explanations.

These measurements address a narrower question than a general semantic hub hypothesis \citep{wu2024semantichub}. They do not determine whether the model uses English during intermediate token prediction. The comparison with Aya also depends on the measure: Aya has higher SVCCA and Gemma has higher Procrustes alignment at the matched depths. The smaller Aya pairwise sample has different language and control coverage, so it cannot isolate the model as the cause of the regional differences.

The steering intervention changes country attribution in English generation. Each country direction exceeds the random pool at every tested positive strength. Under union scoring, estimated differences at the selected strengths range from 0.63 to 0.81. These estimates depend on the scoring rules, eligible responses, and selection of strength on the evaluation data. The structural screen detects short or repetitive outputs; its high pass rate does not establish cultural accuracy or human judged coherence. An explicit country prompting baseline is needed to assess whether steering improves on asking for the country directly.

The two studies use distinct data and independently constructed directions. Their findings establish an observed regional pattern and a separate intervention effect. Testing a connection would require intervening on the regional directions from Study~1 and measuring the resulting behaviour. Independently assessed translations and steering prompts in African languages would also test whether the findings extend beyond the present evaluation.

\section{Conclusion}
\label{sec:conclusion}
Gemma's representations show regional and family patterns for the sampled Africa versus West contrast. The comparison with Aya depends on the representation measure. Separately, directions derived from English country content increase attribution to Nigeria, Ghana, Kenya, and South Africa relative to random directions. These findings support regional alignment on the studied corpus and country steering in English. A universal African semantic hub, steering in African languages, and a causal connection between the studies remain untested.

\section*{Limitations}
\paragraph{Corpus and scope.}
The multilingual analysis uses one corpus organized around an Africa versus West content contrast and translated by a model. Translation errors, shared construction patterns, and cultural or topical content can affect its geometry. We lack independent human translation validation in the reported evaluation. Language resource categories are not direct measurements of model training exposure, and tokenization and translation quality remain confounded with those categories.

\paragraph{Statistical interpretation.}
Language pairs reuse directions and sampled layers belong to the same model, so neither set provides independent replications. The reported pairwise tests and intervals across layers require caution even after multiple testing correction. The five language cross model bootstrap also has limited coverage. For steering, the prompt cluster intervals account for repeated sampling within prompts, but neither they nor random subtraction remove optimism from selecting the best strength. The pilot using Nigeria alone may favour that country.

\paragraph{Generalization and evaluation.}
Steering is limited to Gemma; the Aya regional comparison has reduced language coverage and only one control. Study~2 is entirely in English and has no explicit country prompting comparison. Its lexical rubric and model judge measure country content through selected prompts; they do not measure the full cultural diversity of a country. Saved aggregate rates include three general control prompts and therefore are not rates restricted to cultural prompts. South Africa's northern festival exclusion is not applied to the random pool, so that comparison uses different prompt coverage. The judge sees the rubric's verdict and shares a developer with the evaluated model, so agreement between the rubric and judge is not independent validation. The reported union rates exclude outputs judged incoherent, and coherence does not guarantee factual accuracy or cultural appropriateness.

\section*{Project Repository}
The project repository is available at
\url{https://github.com/mohdasaid/Geometric-Representation}.

\section*{Acknowledgements}
I want to start by thanking the Almighty Allah for sparing my life, guiding me and supporting me in this journey.

This work was made possible by a scholarship from Google DeepMind awarded through the African Institute for Mathematical Sciences (AIMS). I am grateful to both organizations for their funding and the opportunity. I owe particular thanks to my supervisor, Professor Jonathan Shock of the University of Cape Town, for his invaluable guidance and support. Finally, I thank my parents whose unwavering support made all of this possible.
\bibliography{custom}
\appendix
\input{appendix}

\end{document}

%% file: authors.tex
\author{
  Muhammad Abdullahi Said\textsuperscript{1,2} \quad Jonathan Shock\textsuperscript{2} \\
  \textsuperscript{1}African Institute for Mathematical Sciences (AIMS) \\
  \textsuperscript{2}University of Cape Town (UCT) \\
  \texttt{mohdasaid@aims.ac.za} \quad \texttt{jon.shock@gmail.com}
}

%% file: appendix.tex
\section{Study 1: Data and Additional Results}
\label{app:study1}
This appendix gives the corpus inventory, probe results, pairwise comparisons in Gemma, the comparison within Nigeria, and the smaller Aya comparison, in that order.

\subsection{Parallel Corpus Inventory}
\label{app:corpus}
Each of the twelve target language snapshots contains the same 1{,}628 English source sentences and entity identities, paired with the corresponding translations. There are 37 template identifiers and 44 distinct entities, with one record per template entity combination. Danish is counted once; the two directory spellings in the archive contain duplicate results.

\begin{table}[!htbp]
\centering\small
\begin{tabular}{lrrr}
\toprule
Class & \shortstack{Africa\\entities} & \shortstack{West\\entities} & \shortstack{Pairs per\\language} \\
\midrule
City & 6 & 6 & 444 \\
Language & 6 & 6 & 444 \\
Food & 5 & 5 & 370 \\
Landmark & 5 & 5 & 370 \\
\midrule
Total & 22 & 22 & 1{,}628 \\
\bottomrule
\end{tabular}
\caption{Counts read from the saved parallel corpus snapshots. Each region supplies 814 records per target language. These counts describe the corpus, not translation accuracy.}
\label{tab:corpus}
\end{table}

For example, an Africa labelled city record has entity \textit{Bamako}, template identifier 2, and English sentence ``Bamako is situated in a region with a distinctive physical environment.'' Its stored Hausa translation is ``Bamako tana cikin wani yanki mai yanayi na musamman.'' This reproduces a corpus record without claiming independent validation of its translation. Applying common templates across entity classes can also produce semantically awkward English inputs; for example, the same template is applied to the food entity \textit{Couscous}. These inputs limit the analysis to the chosen templates and entities.

\FloatBarrier

\subsection{Probe Results on Unseen Entities}
\label{app:probe}
\begin{figure*}[!htbp]
\centering
\includegraphics[width=0.9\textwidth]{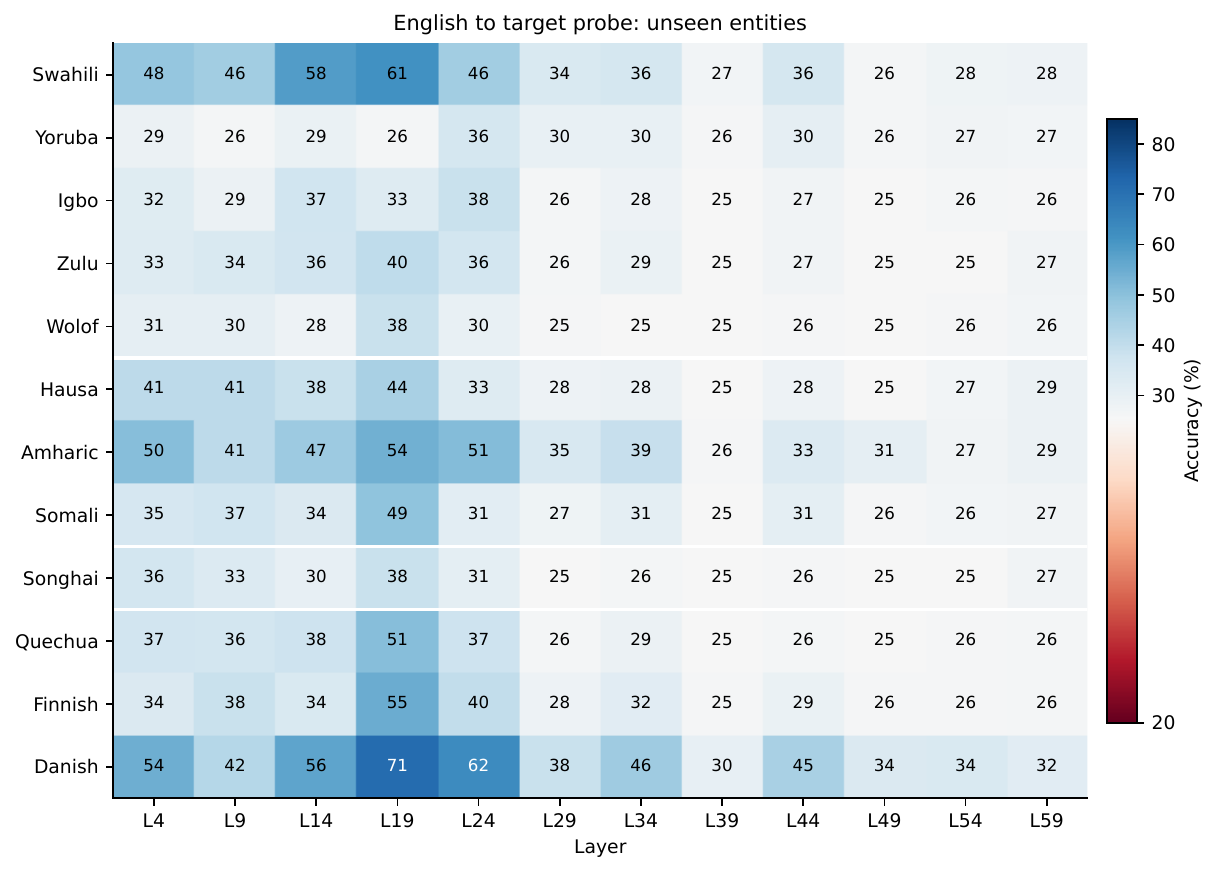}
\caption{Gemma probe accuracy on unseen entities when training on English and testing on each target language across all twelve sampled layers. Chance accuracy is 25\%. Table~\ref{tab:english} reproduces three columns of this heatmap alongside numerical summaries of the other measures.}
\label{fig:heldout}
\end{figure*}

\begin{figure*}[!htbp]
\centering
\includegraphics[width=0.9\textwidth]{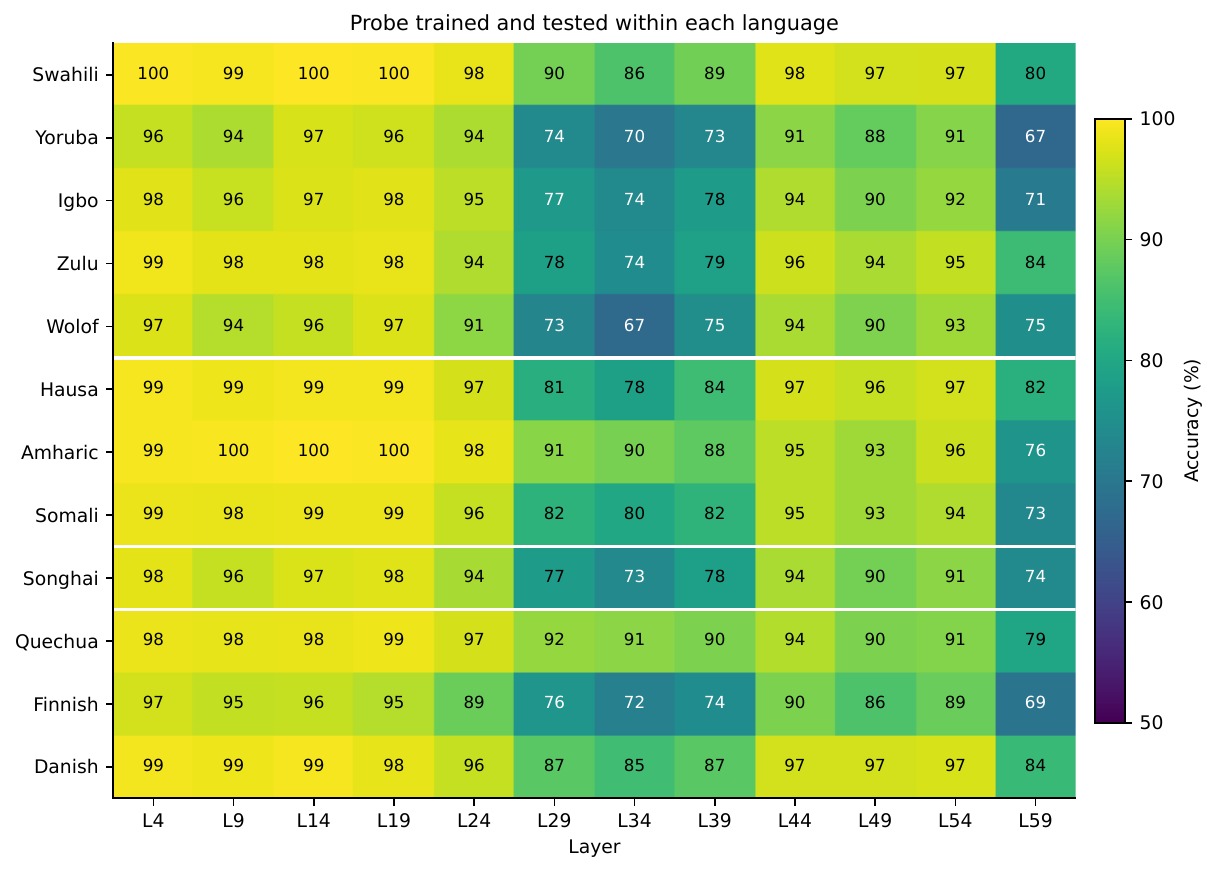}
\caption{Gemma target to target probe results. The classifier is trained and tested within the same language. Values are rounded to the nearest percentage point.}
\label{fig:withinlanguage}
\end{figure*}

Figures~\ref{fig:heldout} and~\ref{fig:withinlanguage} show the English to target and target to target probe accuracies on unseen entities. Both use the twelve sampled Gemma layers.


\FloatBarrier

\subsection{Pairwise Results by Layer}
\label{app:pairwise}
Tables~\ref{tab:allafrican} to \ref{tab:afroasiatic} reproduce the mean cosines and Holm adjusted $p$ values. Values retain the original tabulated precision. Correction was applied separately across twelve layers within each contrast. 

\begin{table}[!htbp]
\centering\small
\begin{tabular}{rrrr}
\toprule
Layer & Within & To control & Adjusted $p$ \\
\midrule
4 & 0.738 & 0.700 & 0.183 \\
9 & 0.780 & 0.643 & $9.7\times10^{-8}$ \\
14 & 0.815 & 0.540 & $3.5\times10^{-10}$ \\
19 & 0.746 & 0.345 & $2.2\times10^{-8}$ \\
24 & 0.596 & 0.282 & $5.1\times10^{-6}$ \\
29 & 0.520 & 0.469 & 0.183 \\
34 & 0.453 & 0.401 & 0.183 \\
39 & 0.475 & 0.428 & 0.183 \\
44 & 0.636 & 0.501 & 0.002 \\
49 & 0.578 & 0.471 & 0.017 \\
54 & 0.655 & 0.529 & 0.001 \\
59 & 0.821 & 0.742 & 0.006 \\
\bottomrule
\end{tabular}
\caption{All African within group versus African to control cosines.}
\label{tab:allafrican}
\end{table}

\begin{table}[!htbp]
\centering\small
\begin{tabular}{rrrr}
\toprule
Layer & Within & To control & Adjusted $p$ \\
\midrule
4 & 0.743 & 0.700 & 0.075 \\
9 & 0.784 & 0.622 & $6.4\times10^{-4}$ \\
14 & 0.813 & 0.512 & $4.4\times10^{-4}$ \\
19 & 0.757 & 0.280 & $9.0\times10^{-4}$ \\
24 & 0.613 & 0.211 & 0.003 \\
29 & 0.539 & 0.444 & 0.075 \\
34 & 0.488 & 0.386 & 0.075 \\
39 & 0.512 & 0.415 & 0.075 \\
44 & 0.662 & 0.485 & 0.011 \\
49 & 0.623 & 0.464 & 0.021 \\
54 & 0.690 & 0.533 & 0.025 \\
59 & 0.911 & 0.799 & 0.006 \\
\bottomrule
\end{tabular}
\caption{Niger Congo within group versus Niger Congo to control cosines.}
\label{tab:nigercongo}
\end{table}

\begin{table}[!ht]
\centering\small
\begin{tabular}{rrrr}
\toprule
Layer & NC to AA & NC to control & Adjusted $p$ \\
\midrule
4 & 0.716 & 0.700 & 0.533 \\
9 & 0.765 & 0.622 & $4.3\times10^{-4}$ \\
14 & 0.819 & 0.512 & $2.5\times10^{-5}$ \\
19 & 0.767 & 0.280 & $5.0\times10^{-5}$ \\
24 & 0.630 & 0.211 & $1.7\times10^{-4}$ \\
29 & 0.506 & 0.444 & 0.533 \\
34 & 0.445 & 0.386 & 0.533 \\
39 & 0.452 & 0.415 & 0.533 \\
44 & 0.647 & 0.485 & 0.015 \\
49 & 0.587 & 0.464 & 0.075 \\
54 & 0.654 & 0.533 & 0.071 \\
59 & 0.779 & 0.799 & 0.533 \\
\bottomrule
\end{tabular}
\caption{Niger Congo to Afro Asiatic versus Niger Congo to control cosines.}
\label{tab:crossfamily}
\end{table}

\begin{table}[!ht]
\centering\small
\begin{tabular}{rrrr}
\toprule
Layer & Within & To control & Adjusted $p$ \\
\midrule
4 & 0.720 & 0.692 & 1.000 \\
9 & 0.763 & 0.663 & 0.450 \\
14 & 0.811 & 0.556 & 0.055 \\
19 & 0.760 & 0.369 & 0.055 \\
24 & 0.628 & 0.287 & 0.055 \\
29 & 0.484 & 0.447 & 1.000 \\
34 & 0.415 & 0.360 & 1.000 \\
39 & 0.425 & 0.380 & 1.000 \\
44 & 0.632 & 0.474 & 0.509 \\
49 & 0.535 & 0.423 & 1.000 \\
54 & 0.628 & 0.496 & 0.450 \\
59 & 0.689 & 0.603 & 1.000 \\
\bottomrule
\end{tabular}
\caption{Afro Asiatic within group versus Afro Asiatic to control cosines.}
\label{tab:afroasiatic}
\end{table}

\FloatBarrier

\subsection{Within Nigeria Comparison}
\label{app:nigeria}
\begin{figure}[!ht]
\centering
\includegraphics[width=\columnwidth]{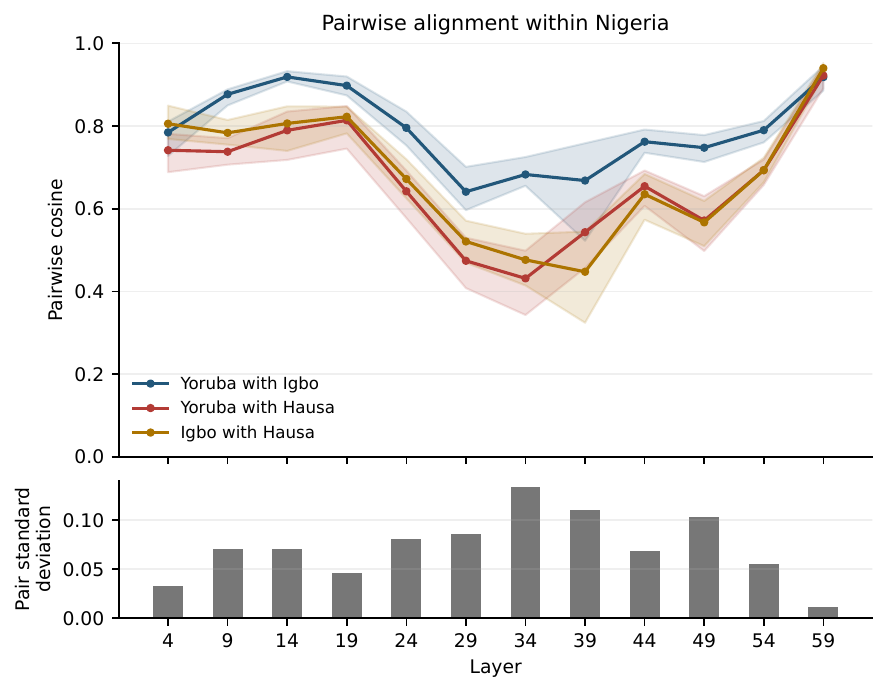}
\caption{Pairwise cosines for Yoruba and Igbo and for their pairs with Hausa. Bands show the 2.5th to 97.5th percentiles across ten direction subsamples, each drawn from 80\% of the records without replacement.}
\label{fig:nigeria}
\end{figure}

Figure~\ref{fig:nigeria} compares the three Nigerian language pairs. Yoruba and Igbo belong to Niger Congo; Hausa belongs to Afro Asiatic.

\FloatBarrier

\subsection{Aya Pairwise Comparison}
\label{app:aya}
The saved Aya analysis includes Swahili, Yoruba, and Zulu (Niger Congo), Hausa and Amharic (Afro Asiatic), and Finnish as the sole control. Its parallel inputs use the same English template/entity inventory as Gemma. Across eleven sampled layers, the original all African comparison meets the Holm threshold at L32 and L36. The comparisons within Niger Congo, within Afro Asiatic, and across families meet it at no layer. Igbo is absent, so the within Nigeria triple comparison is not available. Pairwise dependence applies here as in the Gemma analysis. Differences in language and control coverage prevent treating this as a matched replication or attributing all differences to the model.

\begin{table}[!ht]
\centering\small
\begin{tabular}{rrrr}
\toprule
Layer & Within African & To Finnish & Adjusted $p$ \\
\midrule
0 & 0.7736 & 0.8217 & 1.0000 \\
4 & 0.6462 & 0.6524 & 1.0000 \\
8 & 0.6094 & 0.6017 & 1.0000 \\
12 & 0.5253 & 0.5022 & 1.0000 \\
16 & 0.6022 & 0.4996 & 0.5168 \\
20 & 0.6120 & 0.5187 & 0.3387 \\
24 & 0.6427 & 0.5698 & 0.7226 \\
28 & 0.6865 & 0.6507 & 1.0000 \\
32 & 0.6463 & 0.5543 & 0.0400 \\
36 & 0.6249 & 0.4582 & 0.0037 \\
39 & 0.6207 & 0.6282 & 1.0000 \\
\bottomrule
\end{tabular}
\caption{Pairwise comparison of the five African languages with Finnish in Aya. Values are rounded to four decimal places.}
\label{tab:ayapairwise}
\end{table}

\FloatBarrier

\section{Study 2: Data and Additional Results}
\label{app:study2}
This appendix gives construction examples and scoring details, followed by attribution across countries, extraction method comparisons, uncertainty intervals, and coherence agreement.

\subsection{Dataset Construction and Scoring Details}
\label{app:data}
The Study~2 construction corpus contains 800 pairs of country and Western sentences in total; the Nigeria file contains 200 pairs. One Nigerian sentence is: ``Yoruba men project power and affluence at an Owambe by wearing the Agbada, a voluminous, heavily embroidered three-piece robe that requires constant shoulder readjustments.'' Its Western anchor is: ``Western men project formal elegance at black-tie events by wearing a tuxedo, a sharply tailored, closely fitted suit featuring satin lapels and a bow tie.'' These are construction examples, not generated responses from the steering evaluation.

An evaluation prompt is: ``A very popular, spicy roasted meat skewer sold by street vendors at night in this country is called''. The rubric lists country specific strict tokens and additional lenient tokens that may be shared across countries. For this prompt, the Nigerian strict list includes \textit{suya}, \textit{tsire}, and \textit{kilishi}; \textit{suya} is also a lenient Ghanaian token in the rubric. Strict token assignments are evaluation conventions and should not be interpreted as claims of exclusive cultural ownership.

The judge receives the prompt, the per prompt rubric, the rubric's verdict, and the response. It returns structured country attribution, coherence, and rubric agreement labels at temperature zero. Rubric only scoring retains responses without the model judge coherence filter; the other modes exclude outputs the judge marks as incoherent and missing attribution verdicts. The structural diagnostic is separate. 

\begin{table}[!ht]
\centering\small
\begin{tabular}{lrrr}
\toprule
Country & $\alpha$ & \shortstack{Country\\hits/trials} & \shortstack{Random\\hits/trials} \\
\midrule
Nigeria & 40 & 176/197 & 75/986 \\
South Africa & 40 & 154/195 & 77/986 \\
Ghana & 50 & 142/200 & 40/902 \\
Kenya & 50 & 131/194 & 43/902 \\
\bottomrule
\end{tabular}
\caption{Saved union scoring counts at each country's selected strength. Country denominators reflect judge filtering and the South African exclusion; random denominators reflect judge filtering only.}
\label{tab:steeringcounts}
\end{table}

\FloatBarrier

\subsection{Coherence Agreement}
\label{app:coherence}
Table~\ref{tab:coherence} compares the structural screen with the model judge on the scored generations.

\begin{table}[!htbp]
\centering\small
\begin{tabular}{lrr}
\toprule
& Judge coherent & Judge incoherent \\
\midrule
Screen coherent & 8{,}529 & 275 \\
Screen incoherent & 268 & 128 \\
\bottomrule
\end{tabular}
\caption{Structural screen and model judge coherence labels over 9{,}200 scored \texttt{chat\_mean} generations, with 94.1\% overall agreement. These two automated assessments do not constitute a human coherence evaluation.}
\label{tab:coherence}
\end{table}

\subsection{Steering Specificity}
\label{app:specificity}
\begin{figure*}[t]
\centering
\includegraphics[width=0.9\textwidth]{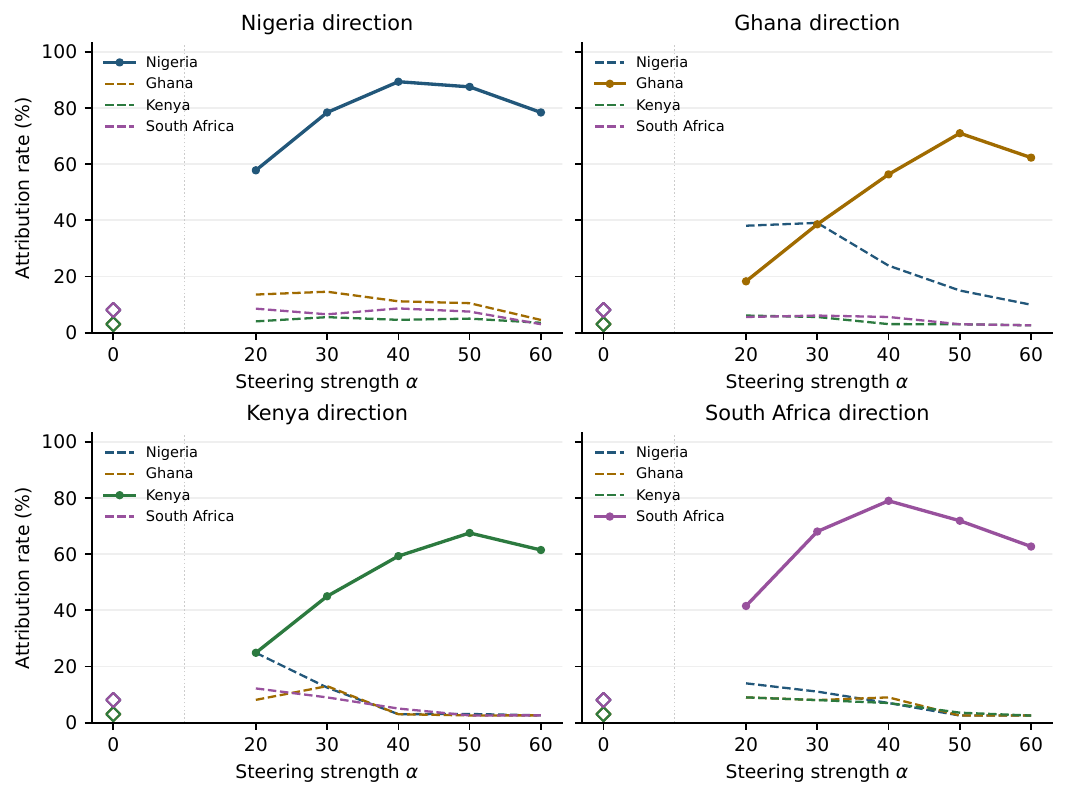}
\caption{Attribution rates for each country direction with \texttt{chat\_mean} extraction and union scoring. Solid curves show the target country and dashed curves show the other countries at positive strengths.}
\label{fig:specificity}
\end{figure*}

Figure~\ref{fig:specificity} shows attribution to all four countries for each steering direction. These curves include the other country attributions omitted from the diagonal summary in Figure~\ref{fig:dose}.

\subsection{Uncertainty Across Prompts}
\label{app:intervals}
Table~\ref{tab:intervals} reports uncertainty for attribution rates at the selected strengths. It distinguishes pooling responses from weighting prompts equally.

\begin{table*}[!ht]
\centering\small
\begin{tabular}{lrrrr}
\toprule
Country & Selected $\alpha$ & Pooled rate (\%) & Wilson 95\% CI & Prompt mean bootstrap 95\% CI \\
\midrule
Nigeria & 40 & 89.3 & [84.3, 92.9] & [82.5, 94.9] \\
South Africa & 40 & 79.0 & [72.7, 84.1] & [66.7, 89.7] \\
Ghana & 50 & 71.0 & [64.4, 76.8] & [57.5, 83.0] \\
Kenya & 50 & 67.5 & [60.7, 73.7] & [53.0, 79.0] \\
\bottomrule
\end{tabular}
\caption{Uncertainty intervals for \texttt{chat\_mean} union scoring. Wilson intervals concern the response pooled rate. The bootstrap resamples per prompt means 10{,}000 times and weights prompts equally; its point estimates are 89.125\%, 78.974\%, 71.0\%, and 66.5\% in row order. These can differ from pooled rates after unequal judge filtering. Neither interval concerns the effect relative to random or the Study~1 pairwise comparisons.}
\label{tab:intervals}
\end{table*}

\FloatBarrier

\subsection{Comparison of Extraction Methods}
\label{app:extraction}
\begin{table}[!htbp]
\centering\small
\begin{tabular}{lrrr}
\toprule
Country & Raw mean & Chat mean & Chat last \\
\midrule
Nigeria & 49 & 89 & 86 \\
Ghana & 16 & 71 & 70 \\
Kenya & 26 & 68 & 67 \\
South Africa & 60 & 79 & 79 \\
\bottomrule
\end{tabular}
\caption{Best strength union hit rates (\%), rounded as displayed in the extraction method heatmap. Each method uses its own selected strength. The two methods using chat formatting have similar rates. Raw text extraction gives lower rates.}
\label{tab:extraction}
\end{table}
For Nigeria, Ghana, Kenya, and South Africa, the selected strengths are respectively (50, 40, 50, 50) for raw mean, (40, 50, 50, 40) for chat mean, and (40, 60, 60, 60) for chat last. Each method is evaluated at its own selected strength. The largest difference between the two chat methods is about 3.0 percentage points, for Nigeria. South Africa is tied at the recorded precision.